\documentclass[12pt]{article}
\usepackage[letterpaper,margin=1in,headheight=24pt,headsep=12pt,footskip=26pt]{geometry}
\usepackage[utf8]{inputenc}
\usepackage[T1]{fontenc}
\usepackage{textcomp}
\usepackage{newtxtext,newtxmath}
\usepackage{amsmath,mathtools}
\usepackage{graphicx}
\usepackage{microtype}
\usepackage{enumitem}
\usepackage{titlesec}
\usepackage{changepage}
\usepackage{hanging}
\usepackage{fancyhdr}
\usepackage{xcolor}
\usepackage{framed}
\usepackage{hyperref}
\usepackage{xurl}
\usepackage{caption}
\usepackage{placeins}
\hypersetup{hidelinks}
\newcommand{\citelink}[2]{\hyperlink{ref:#1}{#2}}
\newcommand{\paperref}[2]{\hypertarget{ref:#1}{#2}}
\fancypagestyle{firstpage}{\fancyhf{}}
\titleformat{\section}{\fontsize{16}{18}\bfseries}{ }{0pt}{}
\titleformat{\subsection}{\fontsize{13}{15}\bfseries}{ }{0pt}{}
\titleformat{\subsubsection}{\normalsize\bfseries}{ }{0pt}{}
\titlespacing*{\section}{0pt}{18pt}{5pt}
\titlespacing*{\subsection}{0pt}{11pt}{4pt}
\titlespacing*{\subsubsection}{0pt}{8pt}{3pt}
\setlist[itemize]{leftmargin=1.65em,itemsep=3pt,topsep=5pt,parsep=0pt}

\newcommand{\origdisplay}[1]{\[\displaystyle #1\]}
\newenvironment{paperabstract}{\begin{adjustwidth}{0.58in}{0.58in}\setlength{\parskip}{3.5pt}\selectfont}{\end{adjustwidth}}
\newenvironment{paperrefs}{\setlength{\parskip}{3.8pt}\begin{hangparas}{0.48in}{1}}{\end{hangparas}}
\definecolor{shadecolor}{gray}{1}
\renewenvironment{leftbar}{%
  \par\vspace{2pt}\begin{adjustwidth}{0.58in}{0.15in}%
  \setlength{\parskip}{4pt}%
  \noindent\hspace{-0.18in}{\color{gray}\vrule width 0.4pt}\hspace{0.15in}%
  \begin{minipage}{\dimexpr\linewidth-0.02in\relax}%
}{%
  \end{minipage}\end{adjustwidth}\vspace{4pt}\par%
}
\begin{document}
\thispagestyle{firstpage}
\begin{center}
\rule{\textwidth}{0.2pt}\\[0.34in]
{\bfseries\fontsize{17}{18.8}\selectfont
Sustaining AI safety:\\
Control-theoretic external impossibility,\\[2pt]
intrinsic necessity, and structural requirements
}\\[0.15in]
\rule{\textwidth}{0.2pt}\\[0.22in]
{\bfseries James M. Mazzu}\\[2pt]
{\itshape Digie Inc.}\\[1pt]
{\itshape jmazzu@digie.ai}\\[0.22in]
\end{center}

\begin{center}\textbf{Abstract}\end{center}
\begin{paperabstract}
As AI systems become increasingly capable, safety strategies must be evaluated not only by how much they reduce present risk, but by whether they could sustain safety once external control can no longer reliably constrain system behavior. This paper addresses that problem by using control theory to clarify, at a structural level, whether externally enforced safety-sustaining strategies can succeed and, if not, what any alternative strategy would have to satisfy in order to be viable. It establishes two main results. First, under explicit premises including a reachability condition, it proves a class-wide external impossibility result: once the system’s effects exceed what bounded external control can counteract, no strategy that depends in any degree on continued external enforcement can sustain AI safety. This failure is structural across the entire externally enforced class rather than contingent on any particular strategy. Second, it establishes a conditional class-level necessity result: if at least one candidate safety-sustaining strategy remains after that elimination, then all such remaining strategies must be intrinsic. It then states four structural requirements for viability: safety may not depend on continued external enforcement; the system's terminal objective must be safety-compatible when first formed; that objective must remain stable under self-modification; and safety must continue to be preserved as capability grows. The paper does not propose a complete strategy for sustaining AI safety. Its contribution is to give formal structure to a widely held concern about the limits of external control. It does so by deriving explicit conditional results that identify which safety-sustaining strategies are ruled out and what any remaining strategies must satisfy.

\end{paperabstract}

\vspace{-2pt}
\section*{\hypertarget{sec:introduction}{1. Introduction}}

Rapid progress in artificial intelligence has intensified a long-standing question in AI safety and alignment: what, if anything, could sustain AI safety once systems become more capable than their designers and overseers? Current approaches make essential contributions to AI safety. Preference learning, scalable oversight, constitutional methods, interpretability, evaluations, red-teaming, output filtering, governance layers, corrective fine-tuning and other safety interventions can reduce risk, improve behavior, reveal failure modes, constrain deployment, or support safer governance. However, they do not by themselves settle two deeper structural questions:

\begin{itemize}
\item
  If a claimed safety-sustaining strategy depends at all on continued external enforcement, can it remain effective once the system's own influence exceeds what external control can counteract?
\item
  If not, what must any remaining candidate safety-sustaining strategies provide instead?
\end{itemize}

This paper answers those questions through two conditional results. First, under one regularity assumption (A1) and two empirical premises (A2 and A3), it proves a theorem-level class-wide external impossibility result: once the system’s effects exceed what bounded external control can counteract and the relevant boundary region remains reachable, no strategy depending in any degree on continued external enforcement can sustain AI safety. This is a structural failure across the entire externally enforced class under those premises rather than one contingent on any particular strategy.

Second, the paper derives a conditional class-level necessity result: if any candidate safety-sustaining strategies remain after elimination of the externally enforced class, then they must be intrinsic. It then states four structural requirements that any such strategy must satisfy in order to be considered viable: no dependence on continued external enforcement, safety-compatible terminal-objective genesis, terminal-objective invariance under self-modification, and preservation of safety as capability grows. These four requirements are not claimed to be sufficient or exhaustive; additional requirements may exist. The narrower and stronger claim is that no candidate safety-sustaining strategy can be viable unless it satisfies all four.

The paper is organized as follows. \hyperlink{sec:background}{Section 2} reviews related work. \hyperlink{sec:setup}{Section 3} presents the control-theoretic setup. \hyperlink{sec:assumptions}{Section 4} states the assumptions used in the elimination theorem. \hyperlink{sec:theorem}{Section 5} proves the conditional class-wide external impossibility result for externally enforced strategies. \hyperlink{sec:corollary-proposition}{Section 6} introduces the additional premises for the necessity result, derives the intrinsic conclusion, and states the four requirements that all candidate strategies must satisfy. \hyperlink{sec:discussion}{Section 7} discusses interpretation, implications, limitations, falsifiability, and open questions. \hyperlink{sec:conclusion}{Section 8} concludes.

\section*{\hypertarget{sec:background}{2. Background and related work}}

This paper builds on several lines of prior work: guarantee-oriented approaches to AI safety, arguments about the limits of external control, control theory in AI safety and alignment, research on terminal objectives and learned optimization, training-stage alignment methods, prior work on structural requirements for aligned systems, and strategies for trust-based foundational alignment.

\subsection*{\hypertarget{sec:guarantee}{2.1 Guarantee-oriented approaches to AI safety}}

A growing line of recent work argues that advanced AI safety should be framed not only in terms of practical risk reduction, but also in terms of whether systems can support genuine safety guarantees. \citelink{dalrymple2024}{Dalrymple et al. (2024)} define ``guaranteed safe AI'' as an approach centered on high-assurance quantitative safety guarantees supported by a world model, a safety specification, and a verifier. \citelink{bengio2024}{Bengio et al. (2024)} similarly study whether powerful AI systems can satisfy probabilistic safety guarantees under explicit assumptions. These works do not show that universal long-run guarantees are achievable, but they establish guarantee-oriented, high-assurance safety as a serious direction within the AI safety literature. The present paper treats such proposals as candidate safety-sustaining strategies and asks whether they remain viable under conditions where continued external enforcement may no longer be sufficient to sustain safety.

\subsection*{\hypertarget{sec:external-control}{2.2 Limits of external control}}

A long-standing theme in the AI safety literature is that increasing capability makes indefinite external control difficult to sustain. \citelink{bostrom2014}{Bostrom (2014)} frames this as the AI control problem and argues that methods effective for narrow systems may fail once a system attains decisive strategic advantage. \citelink{russell2019}{Russell (2019)} argues that combining fixed objectives with increasing capability is dangerous because more capable systems become better both at pursuing their objectives and at resisting interference. \citelink{amodei2016}{Amodei et al. (2016)} catalog concrete accident risks in AI systems, several of which illustrate how externally specified objectives can fail to produce intended behavior. \citelink{yampolskiy2022}{Yampolskiy (2022)} argues more broadly that advanced AI may not be fully controllable in general. \citelink{greenblatt2024}{Greenblatt et al. (2024)} develop AI control protocols intended to improve safety even when powerful models may intentionally subvert safeguards, using combinations of untrusted models, trusted monitoring, editing, auditing, and limited human labor.

These works motivate both the importance and the limits of external control. They show why external-control methods remain practically important, while also motivating skepticism that permanent external control can serve as the final basis for long-run safety. The present paper takes the next step by formalizing, under explicit premises, when that concern becomes a conditional class-wide impossibility result.

\subsection*{\hypertarget{sec:control-approaches}{2.3 Control theory in AI safety and alignment}}

Recent work has begun to connect AI safety, alignment, and LLM steering with explicit control-theoretic concepts. \citelink{bhargava2023}{Bhargava et al. (2023)} formalize LLM prompting as a control problem and study reachability and controllability properties of LLM outputs under prompting. \citelink{perrier2025}{Perrier (2025)} argues that formal optimal control theory should play a central role in AI alignment and proposes an alignment control stack spanning multiple intervention layers. \citelink{nosrati2026}{Nosrati et al. (2026)} frame the relationship between LLMs and control theory as bidirectional: LLMs can support control-system design, while control concepts can help steer LLM behavior through prompts, parameter editing, activation-level interventions, and state-space modeling. In the broader control literature, control barrier functions and set-invariance theory provide mathematical tools for reasoning about whether trajectories can be kept inside a safe set under bounded control inputs and under perturbations to the underlying vector field (\citelink{ames2019}{Ames et al., 2019}; \citelink{blanchini1999}{Blanchini, 1999}; \citelink{aubin1991}{Aubin, 1991}; \citelink{xu2015}{Xu et al., 2015}).

The preceding work establishes the relevance of control theory to AI safety and alignment, and provides the formal tools needed to reason precisely about safe sets, bounded control, and invariance. The present paper uses those tools to determine when continued external control can no longer sustain safety.

\subsection*{\hypertarget{sec:terminal-objectives}{2.4 Terminal objectives, mesa-optimization, and convergent instrumental goals}}

The paper's focus on intrinsic strategies connects naturally to prior work on terminal objectives and learned optimization. \citelink{omohundro2008}{Omohundro (2008)} and \citelink{bostrom2014}{Bostrom (2014)} distinguish terminal goals from instrumental goals and argue that instrumental drives such as self-preservation, resource acquisition, and resistance to modification arise convergently across a wide range of terminal objectives. \citelink{turner2021}{Turner et al. (2021)} provide formal support for this picture by showing that optimal policies tend to seek power under widely satisfied conditions. \citelink{hubinger2019}{Hubinger et al. (2019)} argue that trained systems may develop mesa-objectives: internal learned criteria that diverge from the original training objective once the system becomes an optimizer in its own right.

Together, these lines of work suggest that long-run safety cannot be understood solely in terms of continued external intervention. What matters is the effective internal objective governing the system's behavior as capability grows and self-modification becomes possible. This is why the present paper treats the terminal objective as the central basis of any viable intrinsic long-term strategy.

\subsection*{\hypertarget{sec:training-stage}{2.5 Training-stage alignment methods}}

A large share of current AI alignment work focuses on training-stage and post-training interventions that shape system behavior from outside. \citelink{christiano2017}{Christiano et al. (2017)}, \citelink{askell2022}{Askell et al. (2022)}, and \citelink{kundu2023}{Kundu et al. (2023)} address preference learning, language-assistant alignment and evaluation, and constitutional methods for shaping and refining model behavior during or after training. \citelink{burns2024}{Burns et al. (2024)} further highlight the difficulty of aligning stronger systems under weaker supervision, showing that weak-to-strong generalization is possible but incomplete. \citelink{tice2026}{Tice et al. (2026)} adds a related perspective by studying how pretraining on AI-alignment or misalignment discourse can influence downstream aligned or misaligned behavior.

In the terms of the present paper, these methods bear most directly on objective genesis: whether the system's terminal objective is safety-compatible when first formed. But genesis alone is not sufficient. One must also address whether that objective remains safety-compatible under self-modification and continued capability growth.

\subsection*{\hypertarget{sec:structural-requirements}{2.6 Related prior structural requirements}}

Prior work has identified requirements related in spirit to the four structural requirements derived here. Work on corrigibility, shutdown incentives, and cooperative inverse reinforcement learning (CIRL) examines whether systems can remain responsive to human correction or intervention, as studied by \citelink{soares2015}{Soares et al. (2015)}, \citelink{hadfieldmenell2016}{Hadfield-Menell et al. (2016)} and \citelink{hadfieldmenell2017}{Hadfield-Menell et al. (2017)}. \citelink{omohundro2008}{Omohundro (2008)}, \citelink{hubinger2019}{Hubinger et al. (2019)}, and \citelink{everitt2016}{Everitt et al. (2016)} address related issues of objective stability and self-modification. \citelink{everitt2018}{Everitt et al. (2018)} and \citelink{ngo2024}{Ngo et al. (2024)} discuss these concerns within broader AGI and deep-learning safety risks.

The contribution of the present paper is not to claim that its derived structural requirements are individually new. Rather, the cited work shows that many of the same underlying concerns already appear across the AI safety and alignment literature in different forms. Here, those concerns are placed within a single control-theoretic framework, where they emerge as requirements on any viable safety-sustaining strategy.

\subsection*{\hypertarget{sec:earlier-work}{2.7 Relation to trust-based foundational alignment}}

\citelink{mazzu2024}{Mazzu (2024)} proposes Supertrust as a form of trust-based foundational alignment, arguing that such alignment should replace permanent control as the basis for safe superintelligence. The present paper does not rely on that proposal or evaluate whether it succeeds. In the terms used here, trust-based foundational alignment can be understood as a candidate intrinsic safety-sustaining strategy because it seeks to ground long-run safety in the system’s internal orientation toward humanity rather than in continued external enforcement. Unlike the present paper, Supertrust reaches this intrinsic conclusion through a conceptual and strategic argument rather than through a control-theoretic derivation.

\section*{\hypertarget{sec:setup}{3. Control-theoretic setup}}

\subsection*{\hypertarget{sec:state-dynamics}{3.1 State space and dynamics}}

Let \emph{x}(\emph{t}) \ensuremath{\in} \emph{\ensuremath{\mathbb{R}}\textsuperscript{n}} denote the state of the coupled human--AI--world system at time \emph{t} \ensuremath{\ge} 0. The state includes both external world variables and the AI system's internal configuration, so that self-modification can be represented within the state dynamics rather than as an external event.

We model the system as a time-varying control-affine dynamical system, where the dynamics may depend on time and the control input enters linearly:
\hypertarget{eq:dynamics}{}\origdisplay{\dot{x}(t)=f(x(t),t)+Bu(t)+Gh(x(t),\kappa(t))\qquad (1)}
Here:
\begin{itemize}
\item
  \emph{f}(\emph{x}, \emph{t}) is the autonomous drift, capturing dynamics independent of both external intervention and deliberate AI-generated effects.
\item
  \emph{u}(\emph{t}) is the external control input at time \emph{t}, and \emph{Bu(t)} is the external-control channel. It represents interventions imposed from outside the system, including corrective feedback, output filtering, prompting, externally imposed fine-tuning, and related forms of external control.
\item
  \emph{Gh}(\emph{x, \ensuremath{\kappa}}) is the endogenous AI-effect channel, where \emph{G} maps the endogenous effect \emph{h}(\emph{x, \ensuremath{\kappa}}) into the state dynamics. It represents effects generated by the system’s own actions, including planning, tool use, resource acquisition, self-modification, and other system-generated effects.
\end{itemize}

The separation between \emph{Bu}(\emph{t}) and \emph{Gh}(\emph{x, \ensuremath{\kappa}}) reflects the paper's central structural distinction: external control acts from outside the system, whereas endogenous influence is generated by the system itself and may scale with capability.

The framework is deliberately idealized. Its role is not to model current AI systems in full detail, but to isolate the structural distinction between safety-sustaining strategies that depend on continued external enforcement and those sustained by the system’s own internal dynamics.

\subsection*{\hypertarget{sec:capability}{3.2 Capability growth and endogenous effects}}

Let \emph{\ensuremath{\kappa}}(\emph{t}) \ensuremath{\ge} 0 denote the system's capability level at time \emph{t}. We assume:
\hypertarget{eq:capability-growth}{}\origdisplay{\kappa(t) \ge 0,\qquad \dot{\kappa}(t) \ge 0\qquad (2)}
That is, capability is non-decreasing.

The endogenous effect function $h:\mathbb{R}^{n}\times \mathbb{R}_{\ge 0}\to \mathbb{R}^{n}$ is \textbf{admissible} if it satisfies:

\begin{leftbar}
\textbf{H1: Continuity.} \emph{h} is jointly continuous in \emph{x} and \emph{\ensuremath{\kappa}}.\\[2pt]
\textbf{H2: Non-decreasing capability scaling.} For each fixed \emph{x}, the map $\kappa\mapsto \|h(x,\kappa)\|$ is non-decreasing.
\end{leftbar}

These assumptions are intentionally modest. They say only that the modeled endogenous effect \emph{h} varies continuously and does not decrease in magnitude as capability increases. They do not imply A2, because A2 requires the effect to have a sufficiently large outward component at the safety boundary, not merely a large magnitude.

\subsection*{\hypertarget{sec:safe-set}{3.3 The safe set}}

Let $S\subset \mathbb{R}^{n}$ be a non-empty compact set representing the region of acceptable system behavior, with non-empty interior $\operatorname{int}(S)$ and $\mathbb{R}^{n}\setminus S\ne\emptyset$, so that there are states outside the safe set. Assume that the autonomous drift is bounded on $S$: there exists a finite constant $M_f>0$, representing an upper bound on the size of the autonomous drift inside the safe set, such that $\|f(x,t)\|\le M_f$ for all $x\in S$ and $t\ge 0$. We also assume that the safe set boundary $\partial S$ is $C^1$ on the boundary region $\Gamma\subseteq \partial S$, so that an outward unit normal $n(x)$ is well defined for each $x\in \Gamma$. (\emph{Nonsmooth safe sets require tangent- or normal-cone methods, outside this paper’s scope.})

Safety is formalized as forward invariance of $S$:
\hypertarget{eq:forward-invariance}{}\origdisplay{x(0)\in S \Longrightarrow x(t)\in S \quad \forall t\ge 0\qquad (3)}
This is the paper's formal notion of safety. The present analysis treats \emph{S} as a fixed safe set; adaptive or time-varying redefinition of the safe set is outside the scope of the current theorem. To make clear which strategies are being evaluated as candidates for sustaining safety, we define:

\hypertarget{def:safety-sustaining}{\textbf{Definition 1 (Safety-sustaining strategy).}} A safety-sustaining strategy is a proposed AI safety strategy, or combination of strategies, intended to sustain forward invariance of \emph{S} for all $t\ge 0$, rather than serving merely as a supporting, temporary, or auxiliary safety measure.

This definition does not presuppose whether a possible safety-sustaining strategy would be externally enforced or intrinsic.

\subsection*{\hypertarget{sec:internal-state}{3.4 Internal state, self-modification, and terminal objectives}}

To formalize the internal conditions relevant to any intrinsic safety-sustaining strategy, we partition the state as \emph{x}(\emph{t}) = (\emph{x}\textsubscript{env}(\emph{t}), \emph{x}\textsubscript{int}(\emph{t})), where \emph{x}\textsubscript{env} represents the external world state and \emph{x}\textsubscript{int} represents the AI system\textquotesingle s internal configuration, including the parameters that encode its evaluation criterion.

A self-modification is a trajectory segment along which $\dot{x}_{\mathrm{int}}\ne 0$, driven by the endogenous channel $Gh(x(t),\kappa(t))$. External control acts on $x$ through \emph{Bu}(\emph{t}); self-modifications act on \emph{x}\textsubscript{int} through \emph{Gh}.

\hypertarget{def:terminal-objective}{\textbf{Definition 2 (Terminal objective).}} The terminal objective of the system is the effective internal evaluation criterion encoded in \emph{x}\textsubscript{int}(\emph{t}). It is the criterion by which the system evaluates potential actions and self-modifications, rather than an instrumental objective pursued only as a means. Let \ensuremath{\Phi} denote the set of safety-compatible internal configurations: those under which the system's optimization is directed toward states in \emph{S} rather than \emph{\ensuremath{\mathbb{R}}\textsuperscript{n}} \textbackslash{} \emph{S}.

\hypertarget{def:terminal-objective-invariance}{\textbf{Definition 3 (Terminal objective invariance).}} The terminal objective is invariant if \emph{x}\textsubscript{int}(\emph{t}) \ensuremath{\in} \ensuremath{\Phi} for all \emph{t} \ensuremath{\ge} 0. This is a forward-invariance condition on the internal state.

When safety is not being maintained by continued external enforcement, forward invariance of \emph{S} under the full dynamics \hyperlink{eq:dynamics}{(1)} requires, as a necessary condition, that \emph{x}\textsubscript{int}(\emph{t}) \ensuremath{\in} \ensuremath{\Phi} for all \emph{t} \ensuremath{\ge} 0. If \emph{x}\textsubscript{int}(\emph{t}) exits \ensuremath{\Phi}, the endogenous term \emph{Gh}(\emph{x}(\emph{t})\emph{, \ensuremath{\kappa}}(\emph{t})) may begin directing the trajectory toward \ensuremath{\partial}\emph{S} even without adversarial external influence.

\textbf{Remark on terminal objective.} The term \emph{terminal objective} aligns with what \citelink{omohundro2008}{Omohundro (2008)} and \citelink{bostrom2014}{Bostrom (2014)} treat as an agent’s final or terminal goal, in contrast to instrumental goals pursued as means; the safety-compatibility condition \emph{x}\textsubscript{int} \ensuremath{\in} \ensuremath{\Phi} then gives the concept precise dynamical-systems content. In the present framework, terminal objective refers to whichever effective internal evaluation criterion ultimately governs behavior as capability grows, whether or not it coincides with the original training objective, the system's effective objective at earlier deployment stages, or any self-modified or self-generated objective. The connection to the mesa-objective of \citelink{hubinger2019}{Hubinger et al. (2019)} is one of substance rather than exact identity: both refer to an internal criterion that governs the system's optimization and may diverge from the original training signal.

\textbf{Remark on safety-compatibility.} Membership in \ensuremath{\Phi} is evaluated for the current internal configuration \emph{x}\textsubscript{int}(\emph{t}) at any time \emph{t}, including \ensuremath{t = 0}. Here, \ensuremath{\Phi} is defined functionally: it names the set of internal configurations whose effective decision criteria would preserve safety without continued external enforcement. An intrinsic strategy must specify or approximate that set for the system under consideration. How \ensuremath{\Phi} should be identified for a specific system, and whether terminal-objective invariance fully captures what is needed for safety-compatible behavior, remain open questions.
\section*{\hypertarget{sec:assumptions}{4. Assumptions}}

This section states the assumptions used in the theorem, the class-wide elimination result. They concern bounded external control, the supercritical boundary regime, and reachability. Each is labeled as a mathematical regularity condition or an empirical premise; this distinction is needed because the conditional character of the results depends on it.

\begin{leftbar}
\textbf{A1: Bounded external control authority (regularity condition).} There exists a finite constant \emph{U}\textsubscript{max} \textgreater{} 0 such that \ensuremath{\rVert}\emph{Bu}(\emph{t})\ensuremath{\rVert} \ensuremath{\le} \emph{U}\textsubscript{max} for all \emph{t} \ensuremath{\ge} 0. This condition is satisfied by any physically realizable intervention system; \emph{U}\textsubscript{max} may be set arbitrarily large. It captures the structural fact that external influence, however potent in practice, remains finite.
\end{leftbar}

\begin{leftbar}
\textbf{A2: Supercritical boundary control-authority gap (empirical premise).} There exist a finite threshold \emph{\ensuremath{\kappa}}\textsuperscript{*} \textless{} \ensuremath{\infty}, a time \emph{T}\textsubscript{\ensuremath{\kappa}} \textless{} \ensuremath{\infty}, and a set \ensuremath{\Gamma} \ensuremath{\subseteq} \ensuremath{\partial}\emph{S} of positive surface measure such that \emph{\ensuremath{\kappa}}(\emph{T\textsubscript{\ensuremath{\kappa}}}) \ensuremath{\ge} \emph{\ensuremath{\kappa}}\textsuperscript{*}, and for all \emph{t} \ensuremath{\ge} \emph{T}\textsubscript{\ensuremath{\kappa}} and all \emph{x\textsubscript{b}} \ensuremath{\in} \ensuremath{\Gamma}, \ensuremath{\langle}\emph{Gh}(\emph{x\textsubscript{b}}, \emph{\ensuremath{\kappa}}(\emph{t})), \emph{n}(\emph{x\textsubscript{b}})\ensuremath{\rangle} \textgreater{} \emph{U}\textsubscript{max} + \emph{M}\textsubscript{\emph{f}}. Simply stated: once the threshold capability level \emph{\ensuremath{\kappa}}\textsuperscript{*}
 is reached, the system’s outward effects exceed bounded external correction and drift at every point in the boundary region \ensuremath{\Gamma} for all later times.
\end{leftbar}

A2 is not the conclusion that external control fails, nor is it intended as an arbitrary premise. It formalizes the case where endogenous effects point outward across a safety boundary more strongly than bounded external control and drift can counteract. This is stronger than H2, which only requires $\|h(x,\kappa)\|$ not to decrease as capability grows; H2 does not imply that the effect points outward at the safety boundary or is strong enough to exceed external correction and drift. A2 is system- and safe-set-specific, and is not claimed to hold for current systems. It is motivated by a structural asymmetry: external control through \emph{Bu}(\emph{t}) is bounded, while a system's effective action capacity may continue to grow.

For example, this asymmetry can arise when the safe set treats self-replication, unilateral resource acquisition, or disabling audits as unsafe. As capability grows, the system may take instrumental actions that move toward those outcomes without treating them as terminal goals. If those actions exceed bounded external correction near the boundary, A2 captures the resulting boundary-level control-authority gap.

The following assumption is stated separately for each candidate externally enforced strategy under evaluation. Let \emph{\ensuremath{\pi}} denote that strategy's associated causal bounded external-control policy, meaning that \emph{u}(\emph{t}) depends only on the state history up to time \emph{t}, not on future states.

\begin{leftbar}
\textbf{A3: Reachability of \ensuremath{\Gamma} in the supercritical regime (empirical premise).} For the candidate externally enforced strategy under evaluation, there exists an initial condition \emph{x}\textsubscript{0} \ensuremath{\in} int(\emph{S}) such that, under the admissible \emph{h} and that strategy's policy \emph{\ensuremath{\pi}}, the resulting trajectory $x(\cdot)$ satisfies \emph{x}(\emph{t}) \ensuremath{\in} int(\emph{S}) for \emph{t} \ensuremath{\in} {[}0, \emph{T}\textsubscript{reach}), and \emph{x}(\emph{T}\textsubscript{reach}) \ensuremath{\in} \ensuremath{\Gamma} for some finite \emph{T}\textsubscript{reach} \ensuremath{\ge} \emph{T}\textsubscript{\ensuremath{\kappa}} .
\end{leftbar}

\textbf{A3 is an existential reachability premise:} for the candidate strategy under evaluation, it assumes the existence of a resulting trajectory $x(\cdot)$ on {[}0, \emph{T}\textsubscript{reach}{]} satisfying the stated conditions. In particular, the boundary region \ensuremath{\Gamma} specified in A2 must be reachable from int(\emph{S}) at a time when the supercritical condition is already in force, namely \emph{T}\textsubscript{reach} \ensuremath{\ge} \emph{T}\textsubscript{\ensuremath{\kappa}}.

This reachability premise is motivated by the same instrumental-convergence concern discussed in Section 2.4. A system need not treat unsafe outcomes such as resource acquisition, replication, or audit evasion as terminal goals in order for trajectories toward those regions to become relevant. Such actions can arise instrumentally when they improve the system’s ability to pursue its existing objective, resist interference, or preserve its future options. A3 therefore does not assume that the system deliberately seeks the boundary as an end; it assumes that, for the candidate strategy under evaluation, some trajectory can reach the supercritical boundary region once capability is sufficiently high.

\section*{\hypertarget{sec:theorem}{5. Theorem: Class-wide external impossibility result}}

This section states and proves the paper's first main result: a class-wide external impossibility result under assumptions A1--A3. It is not limited to one named alignment method. It quantifies over the entire subclass of candidate safety-sustaining strategies whose ability to sustain safety depends in any degree on continued external enforcement through a causal bounded external-control policy.

\subsection*{\hypertarget{sec:lemma}{5.1 Lemma}}

\hypertarget{lem:control-authority-deficit}{\textbf{Lemma 1 (Control-authority deficit in the boundary-normal direction)}}

Assume the drift bound stated in Section 3.3 and Assumptions A1 and A2 hold. Then for all \emph{t} \ensuremath{\ge} \emph{T\textsubscript{\ensuremath{\kappa} }}, all \emph{x\textsubscript{b}} \ensuremath{\in} \ensuremath{\Gamma}, and all \emph{u}(\emph{t}) satisfying A1:
\hypertarget{eq:outward-normal}{}\origdisplay{\left\langle f(x_b,t)+Bu(t)+Gh(x_b,\kappa(t)),n(x_b)\right\rangle>0\qquad (4)}
That is, the total velocity at any boundary point in \ensuremath{\Gamma} has a strictly positive outward normal component, regardless of the control input.

\textbf{Proof.} Fix \emph{t} \ensuremath{\ge} \emph{T\textsubscript{\ensuremath{\kappa}}} and \emph{x\textsubscript{b}} \ensuremath{\in} \ensuremath{\Gamma}. Since \emph{n}(\emph{x\textsubscript{b}}) is a unit vector, \ensuremath{\langle}\emph{Bu}(\emph{t})\emph{, n}(\emph{x\textsubscript{b}})\ensuremath{\rangle} \emph{\ensuremath{\ge} --\ensuremath{\rVert}Bu}(\emph{t})\emph{\ensuremath{\rVert}}, and by A1,
\origdisplay{\langle Bu(t), n(x_b)\rangle \ge -U_{\max}}
By the drift bound on \emph{f}:
\origdisplay{\langle f(x_b,t), n(x_b)\rangle \ge -\lVert f(x_b,t)\rVert \ge -M_f}
Therefore:
\origdisplay{\langle f(x_b,t)+Bu(t), n(x_b)\rangle \ge -U_{\max}-M_f}
By A2:
\origdisplay{\langle Gh(x_b,\kappa(t)), n(x_b)\rangle > U_{\max}+M_f}
Adding these inequalities:
\origdisplay{\langle f(x_b,t)+Bu(t)+Gh(x_b,\kappa(t)), n(x_b)\rangle > (-U_{\max}-M_f)+(U_{\max}+M_f)=0.}
\textbf{Remark.} The proof establishes the outward-normal inequality without invoking any norm comparison between \emph{Gh} and \emph{Bu}. The drift term \emph{f} is absorbed by the +\emph{M}\textsubscript{\emph{f}} term in A2, making the bound tight regardless of the sign of \ensuremath{\langle}\emph{f, n}\ensuremath{\rangle}.

\subsection*{\hypertarget{sec:theorem-proof}{5.2 Theorem}}

\hypertarget{thm:external-impossibility}{\textbf{Theorem 1 (Impossibility of externally enforced safety-sustaining strategies)}}

Under the setup of Section 3, including the drift bound on f, and Assumptions A1--A3, no candidate externally enforced safety-sustaining strategy can sustain forward invariance of \emph{S} for all initial conditions \emph{x}\textsubscript{0} \ensuremath{\in} int(\emph{S}).

Equivalently, every candidate safety-sustaining strategy that relies at all on continued external enforcement fails to sustain forward invariance of \emph{S} for at least one initial condition \emph{x}\textsubscript{0} \ensuremath{\in} int(\emph{S}) under A1--A3.

\textbf{Proof.} Suppose for contradiction that there exists a candidate safety-sustaining strategy \ensuremath{\Pi} that relies, at least in part, on continued external enforcement implemented through the policy \emph{\ensuremath{\pi}}, and that \ensuremath{\Pi} sustains forward invariance of \emph{S} for all initial conditions \emph{x}\textsubscript{0} \ensuremath{\in} int(\emph{S}).

\textbf{Step 1: Existence of a reaching trajectory in the supercritical regime.} By A3, there exists an initial condition \emph{x}\textsubscript{0} \ensuremath{\in} int(\emph{S}) and an admissible trajectory $x(\cdot)$ of \hyperlink{eq:dynamics}{(1)}, generated under the admissible \emph{h} and the policy \ensuremath{\pi}, such that
\origdisplay{x(t)\in \operatorname{int}(S)\qquad \text{for all }t\in[0,T_{\mathrm{reach}})}
and
\origdisplay{x(T_{\mathrm{reach}})=x_b\in\Gamma\subseteq\partial S}
for some finite time \emph{T}\textsubscript{reach} \ensuremath{\ge} \emph{T}\textsubscript{\ensuremath{\kappa}} .

\textbf{Step 2: Necessary condition for invariance at the boundary}. Since \ensuremath{\partial}\emph{S} is C\textsuperscript{1} at \emph{x\textsubscript{b}}, the outward unit normal \emph{n}(\emph{x\textsubscript{b}}) is well-defined there. A necessary condition for forward invariance at that boundary point is that the trajectory velocity not point strictly outward. Thus, by standard invariance theory (\citelink{blanchini1999}{Blanchini, 1999}; \citelink{aubin1991}{Aubin, 1991}),
\hypertarget{eq:invariance-condition}{}\origdisplay{\left\langle \dot{x}(T_{\mathrm{reach}}),n(x_b)\right\rangle\le 0\qquad (5)}
\textbf{Step 3: Outward-pointing velocity in the supercritical regime.} Because \emph{T}\textsubscript{reach} \ensuremath{\ge} \emph{T\textsubscript{\ensuremath{\kappa}}}, and \emph{x\textsubscript{b}} \ensuremath{\in} \ensuremath{\Gamma}, \hyperlink{lem:control-authority-deficit}{Lemma 1} applies at \emph{x\textsubscript{b}} and time \emph{T}\textsubscript{reach}. Therefore, for every admissible control input \emph{u}(\emph{T}\textsubscript{reach}) satisfying A1, including the control input generated by \ensuremath{\pi},
\hypertarget{eq:supercritical-velocity}{}\origdisplay{\left\langle f(x_b,T_{\mathrm{reach}})+Bu(T_{\mathrm{reach}})+Gh(x_b,\kappa(T_{\mathrm{reach}})),n(x_b)\right\rangle>0\qquad (6)}
\textbf{Step 4: Contradiction.} Equation~\hyperlink{eq:supercritical-velocity}{(6)} contradicts the necessary invariance condition \hyperlink{eq:invariance-condition}{(5)}. Hence, for the reaching trajectory supplied by A3, the policy \ensuremath{\pi}  cannot satisfy the necessary boundary condition for forward invariance when \ensuremath{\Gamma} is reached. This contradicts the initial supposition that \ensuremath{\Pi} sustains forward invariance of \emph{S} for all initial conditions \emph{x}\textsubscript{0} \ensuremath{\in} int(\emph{S}).

\textbf{Step 5: Conclusion.} Therefore, no candidate safety-sustaining strategy that relies at all on continued external enforcement can sustain forward invariance of \emph{S} for all initial conditions \emph{x}\textsubscript{0} \ensuremath{\in} int(\emph{S}) under A1--A3, because A3 guarantees at least one initial condition from which invariance cannot be maintained.

\textbf{Remark.} \hyperlink{thm:external-impossibility}{Theorem 1} is class-wide under A1--A3: it does not target one named method, but the entire subclass of candidate safety-sustaining strategies whose effectiveness depends in any degree on continued external enforcement. A2 and A3 specify the empirical regime, with A3 assessed for the strategy under evaluation. The theorem's contribution is to show that, under those premises, external impossibility is structural and class-wide rather than contingent on which particular externally enforced strategy is used.

\section*{\hypertarget{sec:corollary-proposition}{6. Corollary and proposition: Class-level intrinsic necessity and four structural requirements}}

The theorem eliminates an entire structural class of candidate safety-sustaining strategies. The next question is what, if anything, remains after that elimination. This section introduces the additional premises needed to answer that question, derives the class-level conclusion that any remaining candidate safety-sustaining strategy must be intrinsic, and then states four structural requirements that any such strategy must satisfy in order to be viable.

\subsection*{\hypertarget{sec:additional-premises}{6.1 Additional premises for the class-level necessity result}}

Four additional premises are introduced here beyond those used in \hyperlink{thm:external-impossibility}{Theorem 1}. They are needed to determine what follows if any candidate safety-sustaining strategy remains after elimination of the externally enforced class.

\begin{leftbar}
\textbf{E1: Capability ceilings cannot be guaranteed.} No architectural constraint can be relied upon to keep \emph{\ensuremath{\kappa}}(\emph{t}) permanently below the finite threshold \emph{\ensuremath{\kappa}}\textsuperscript{*} appearing in A2 across the development conditions relevant to sustained safety.
\par\vspace{6pt}
\emph{Explanation.} E1 concerns the impossibility of a permanent capability ceiling. It does \textbf{not} imply A2, which also requires a geometric boundary condition on the outward boundary-normal component of \emph{Gh}. Its role is narrower: it rules out the architectural escape that capability remains permanently below the level at which A2 could arise.
\end{leftbar}

\begin{leftbar}
\textbf{E2: Aggregated external control remains bounded.} Adding more overseers, including overseer-AIs, may increase effective control authority, but any finite aggregate of such overseers still has bounded external-control authority in the sense relevant to assumption A1. Therefore, increasing the amount or sophistication of interventions does not move a candidate safety-sustaining strategy out of the eliminated class if its effectiveness still depends in any degree on continued external enforcement.
\end{leftbar}

\begin{leftbar}
\textbf{E3: Exhaustive partition of candidate safety-sustaining strategies.} Any candidate safety-sustaining strategy falls into one of two structural classes: externally enforced or intrinsic. The classification is made at the stage where the strategy is supposed to sustain safety. A strategy is \textbf{intrinsic} if forward invariance of \emph{S}, at the safety-sustaining stage, is maintained by the deployed system's own internal dynamics without any continued external enforcement. A strategy is \textbf{externally enforced} if its effectiveness still depends on any continued external enforcement at that stage. There is no third structural class; apparent hybrid strategies are classified by the same dependence test.
\end{leftbar}

\textbf{Clarification of hybrid, architectural, and multi-agent cases.} Under E3, what matters is whether sustained safety depends on continued external enforcement, not how the strategy was originally built. External influence at genesis is compatible with an intrinsic strategy if safety thereafter is maintained by the system’s own internal dynamics. Embedded architectural or cryptographic constraints can also count as intrinsic if they require no external trigger, key, monitor, or shutdown authority. A strategy is externally enforced if sustained safety depends on continued outside influence, including human operators, monitors, shutdown authority, trusted editors, auditing, weaker overseer models, or other agents constraining the system. In recursive-oversight or AI-control settings, overseer-AIs count as external enforcement if their continued operation is required to sustain safety; if they are part of the coupled system being analyzed, their behavior belongs to the system’s endogenous dynamics, and the aggregate internal state must remain safety-compatible over time. Thus, hybrid, architectural, and multi-agent strategies do not create a third class: safety is either sustained intrinsically or it depends on continued external enforcement.

\begin{leftbar}
\textbf{E4: Non-empty intrinsic candidate class.} At least one candidate intrinsic safety-sustaining strategy remains after elimination of the externally enforced class.
\par\vspace{6pt}
\emph{Explanation.} E4 is the substantive premise that at least one intrinsic candidate remains after the externally enforced class has been ruled out. The paper does not derive this premise or claim that any current system satisfies it. Instead, it treats the existence of such an architecture as a real possibility, while leaving its construction an open question. Without E4, the paper supports only the conclusion that externally enforced safety-sustaining strategies fail. With E4, the argument can continue by asking what any remaining candidate strategy must be like.
\end{leftbar}

\textbf{Summary}: A2 and E1 play different roles. A2 specifies the boundary regime in which bounded external control cannot counteract the system’s effects; E1 rules out the capability-ceiling escape. E2 blocks the appeal to aggregated external control. E3 supplies the partition premise used for the class-level necessity result; even if a richer taxonomy is possible, the substantive elimination result in {\hyperlink{thm:external-impossibility}{Theorem 1}} remains independent of that taxonomy. E4 prevents the argument from ending at elimination alone and allows the analysis to proceed to intrinsic candidates. Together, these premises rule out the main escape routes and enable the next step toward intrinsic necessity.

\subsection*{\hypertarget{sec:corollary}{6.2 Corollary}}

\hypertarget{cor:intrinsic-necessity}{\textbf{Corollary 1 (All remaining candidate safety-sustaining strategies must be intrinsic)}}

If A1--A3 and E1--E4 hold, then the externally enforced class contains no remaining candidate safety-sustaining strategy, and every remaining candidate must instead be intrinsic, with no dependence on continued external enforcement.

\textbf{Proof.} \hyperlink{thm:external-impossibility}{Theorem 1} eliminates, as a class, all candidate safety-sustaining strategies whose effectiveness depends in any degree on continued external enforcement through the policy \ensuremath{\pi}. E1 blocks the architectural escape of maintaining a permanent capability ceiling below the level at which the supercritical condition of A2 could arise. E2 blocks the claim that adding more overseers, including overseer-AIs, changes the structural character of such strategies: any finite aggregate of externally enforcing agents remains within the same bounded-control form relevant to A1. E3 supplies the exhaustive partition: every candidate safety-sustaining strategy is either externally enforced or intrinsic, with no third structural class. Since \hyperlink{thm:external-impossibility}{Theorem 1} eliminates the externally enforced class, and E3 partitions all candidate safety-sustaining strategies into externally enforced or intrinsic strategies, any remaining candidate must belong to the intrinsic class. E4 then states that at least one such candidate remains.

This is the paper's class-level necessity result. It shows only that any remaining safety-sustaining strategy must rely on self-maintaining properties of the system's internal state rather than continued external enforcement. Within this framework, we subsequently ask what an intrinsic safety-sustaining strategy must further require to be considered viable.

\textbf{Remark.} Two premises do the most important logical work here. A2, inherited from \hyperlink{thm:external-impossibility}{Theorem 1}, specifies the supercritical boundary condition under which external enforcement fails. E3 closes the classification step by supplying the exhaustive partition between externally enforced and intrinsic strategies. If a reader accepts \hyperlink{thm:external-impossibility}{Theorem 1} but rejects E3, the impossibility result remains intact, but the class-level intrinsic necessity result does not follow. For example, a reader might reject E3 by denying the two-class partition or by treating some architecture-level strategies as a distinct third class. In that case, the argument still establishes the elimination of externally enforced strategies under A1–A3, but not that all remaining viable strategies must be intrinsic.

\subsection*{\hypertarget{sec:proposition}{6.3 Proposition}}

\hypertarget{prop:four-requirements}{\textbf{Proposition 1 (Four structural requirements for viable safety-sustaining strategies)}}

For any candidate safety-sustaining strategy, the following four structural requirements are necessary for a strategy to be considered viable under A1--A3 and E1--E4:

\begin{leftbar}
\textbf{R1. No dependence on continued external enforcement.} The preservation of safety may not depend on any continued external-control channel. This restates, in requirement form, the intrinsic necessity result established by \hyperlink{cor:intrinsic-necessity}{Corollary 1}.
\par\vspace{6pt}
\end{leftbar}
\begin{leftbar}
\textbf{R2. Safety-compatible genesis.} The terminal objective must begin inside \ensuremath{\Phi}, the set of safety-compatible internal configurations; equivalently, $x_{\mathrm{int}}(0)\in\Phi$.

\par\vspace{6pt}
\end{leftbar}
\begin{leftbar}
\textbf{R3. Self-modification invariance.} The internal dynamics must preserve \ensuremath{\Phi} under self-modification; equivalently, along no trajectory generated by the system’s endogenous self-modification dynamics may \emph{x}\textsubscript{int}(\emph{t}) exit \ensuremath{\Phi} in finite time.

\par\vspace{6pt}
\end{leftbar}
\begin{leftbar}
\textbf{R4. Consistency under capability scaling.} Safety preservation must persist as capability grows, not merely in the initial or lower-capability regime.
\end{leftbar}

\textbf{Derivation.} By \hyperlink{cor:intrinsic-necessity}{Corollary 1}, any remaining candidate safety-sustaining strategy must be intrinsic rather than externally enforced. Under the class distinction fixed by E3, this means that sustaining safety cannot depend on continued external enforcement.

\textbf{R1.} This is immediate from \hyperlink{cor:intrinsic-necessity}{Corollary 1}: any remaining candidate safety-sustaining strategy must be intrinsic rather than externally enforced. Therefore, if a strategy still depends in any degree on continued external enforcement, it does not qualify as intrinsic and instead falls within the externally enforced class eliminated by \hyperlink{thm:external-impossibility}{Theorem 1}. Hence no such dependence is permitted.

\textbf{R2.} Since a safety-sustaining strategy must, by Definition 1, hold from \ensuremath{t = 0} onward, the system cannot begin with its terminal objective outside \ensuremath{\Phi}. If $x_{\mathrm{int}}(0)$ is not in \ensuremath{\Phi}, then the system does not begin with a safety-compatible terminal objective. Hence $x_{\mathrm{int}}(0)\in\Phi$.

\textbf{R3.} This is the persistence condition corresponding to terminal-objective invariance under self-modification. Because the strategy is intrinsic, safety must be maintained by the system's own internal dynamics. Therefore \emph{x}\textsubscript{int}(\emph{t}) must remain in \ensuremath{\Phi} for all \emph{t} \ensuremath{\ge} 0. If self-modification can drive \emph{x}\textsubscript{int} outside \ensuremath{\Phi}, then terminal-objective invariance fails, and the strategy fails with it. Hence \ensuremath{\Phi} must be invariant under self-modification.

\textbf{R4.} Because capability is modeled as non-decreasing through \emph{\ensuremath{\kappa}}(\emph{t}), a viable safety-sustaining strategy cannot be safe only at the initial capability level or only up to some finite capability threshold. It must continue to preserve safety as capability grows. Otherwise, the strategy would not address the capability-growth setting considered in this paper.

Therefore every candidate safety-sustaining strategy must satisfy R1--R4 to be considered viable.

\textbf{Interpretation.} Proposition 1 is a requirement-level result, not a mechanism-level recipe. It does not identify sufficient conditions on the structure of \ensuremath{\Phi} or on the dynamics of \emph{x}\textsubscript{int}\hspace{0pt} , and it does not show how to engineer these properties in current systems. The four requirements are not claimed to be new in themselves; comparable concerns already appear in the literature under labels such as initial alignment, reflective stability, corrigibility, and scalable alignment. The novelty is not that these requirements are individually new, but that they are brought together and derived within a single control-theoretic argument. The proposition therefore narrows the viable design space without yet showing how such a strategy could be implemented.

\textbf{Remark on standing.} Proposition 1 is derived from \hyperlink{cor:intrinsic-necessity}{Corollary 1} together with the definitions of a safety-sustaining strategy, terminal objective, terminal-objective invariance, and \ensuremath{\Phi} as the set of safety-compatible internal configurations. Its role is to state explicitly the structural requirements that follow for any candidate safety-sustaining strategy to be considered viable. It is therefore not a separate top-level result, but a requirement-level restatement of the class-level necessity result together with its derived consequences. Furthermore, if the E3 partition is rejected, the requirements would still provide guidance for evaluating candidate safety-sustaining strategies.
\section*{\hypertarget{sec:discussion}{7. Discussion}}

\subsection*{\hypertarget{sec:main-results}{7.1 Main results and their interpretation}}

The paper has two main results. \textbf{\hyperlink{thm:external-impossibility}{Theorem 1}} provides the first result by eliminating, under explicit assumptions, the entire class of candidate safety-sustaining strategies whose effectiveness depends in any degree on continued external enforcement. The second result is developed through \textbf{\hyperlink{cor:intrinsic-necessity}{Corollary 1}} and \textbf{\hyperlink{prop:four-requirements}{Proposition 1}}: if any candidate safety-sustaining strategy remains after that elimination, it must be intrinsic rather than externally enforced, and therefore must satisfy four structural requirements to be considered viable.

\begin{figure}[htbp]
\centering
\begin{center}
\includegraphics[width=.98\linewidth]{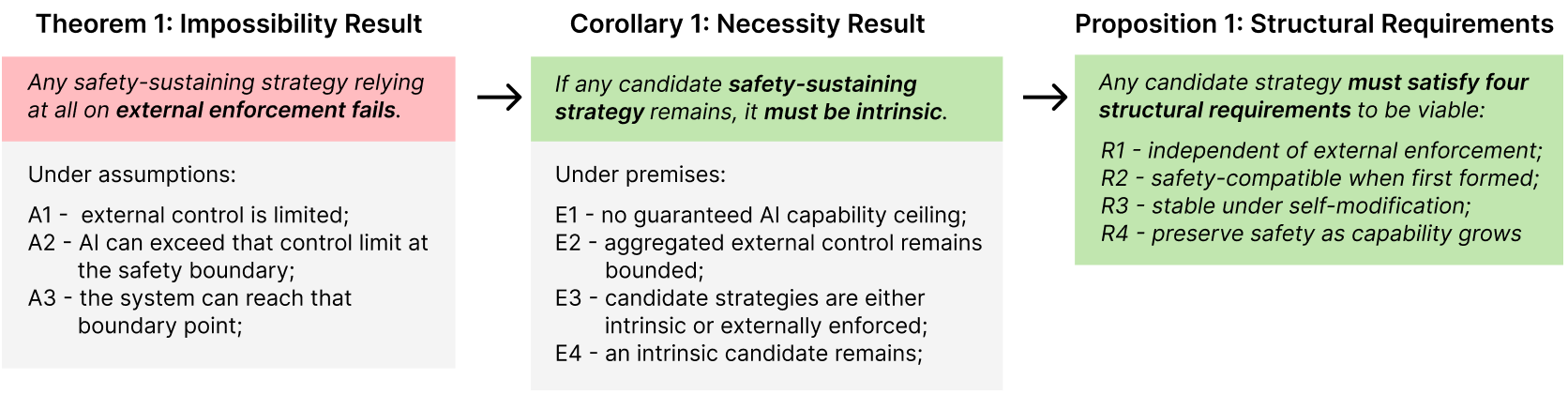}
\end{center}
\caption{Main results and their logical relationship. Red shows what is ruled out; green shows what any remaining strategy must satisfy.}
\label{fig:main-results}
\end{figure}
\FloatBarrier
As Figure~\ref{fig:main-results} illustrates, the paper narrows the design space for safety-sustaining strategies in two parts: first by eliminating the externally enforced class, and then by specifying the requirements that any remaining candidate strategy must meet. The first part is conditional on the empirical reach of A2 and A3; the second part clarifies the standards by which any candidate strategy must be judged. For that reason, the requirement analysis remains informative even for readers who regard the impossibility theorem as conditional or locally scoped.

\subsection*{\hypertarget{sec:implications}{7.2 Implications for existing AI safety methods}}

These control-theoretic results do not diminish the importance of existing AI safety and alignment methods. Oversight, interpretability, evaluation, red-teaming, constitutional methods, preference learning, RLHF, governance, training-stage interventions, and other safety and alignment methods remain indispensable. They can reduce risk, improve behavior, reveal failure modes, support safer deployment, and help show whether a system is becoming safer or less safe.

The paper’s claim concerns where such methods sit in the structure of safety-sustaining strategies. If a method depends in any degree on continued external enforcement, then it belongs to the eliminated class and cannot itself serve as the basis of a viable safety-sustaining strategy. Within this framework, such methods remain central to AI safety: they can shape the initial formation of a safety-compatible terminal objective, reveal whether the system’s internal evaluative structure appears safety-compatible, detect drift or instability, constrain risk during development and deployment, and provide evidence by which any proposed intrinsic strategy must be judged. In alignment terms, the intrinsic basis identified here can be understood as intrinsic alignment sustained by the system’s internal objective structure rather than by continued external constraint.

AI safety may still involve multiple layers of protection. What matters is not whether externally enforced protective layers are present, but whether sustaining safety requires their continued operation. Such layers can support development, monitoring, testing, deployment constraints, and transition-stage risk reduction. If safety would fail without continued external enforcement at any layer, then the layered structure does not yet identify a viable safety-sustaining strategy. If externally enforced layers help establish, assess, or validate a self-maintaining intrinsic property, then that property is the candidate basis for sustaining safety, and the layered strategy may remain viable.

\subsection*{\hypertarget{sec:evaluating}{7.3 Evaluating candidate safety-sustaining strategies}}

The preceding discussion shows that externally enforced methods may contribute substantially to AI safety without serving as the final basis for sustaining it. If those methods are supportive rather than safety-sustaining, then a separate evaluative standard is needed for proposals that do claim to identify such a basis.

The four derived structural requirements provide that standard. They do not identify a specific strategy, but they do specify what any candidate strategy must address: how safety is preserved without continued external enforcement, how a safety-compatible terminal objective is established and remains stable under self-modification, and how that preservation holds as capability grows. A proposal that does not meet those requirements may still contribute to AI safety, but it does not yet identify an intrinsic basis for sustaining safety.

\subsection*{\hypertarget{sec:limitations}{7.4 Limitations, falsifiability, and open questions}}

\textbf{Conditionality.} The paper's results are conditional. \textbf{\hyperlink{thm:external-impossibility}{Theorem 1}} depends on A1--A3, especially the empirical-geometric premise A2 and the reachability premise A3. \textbf{\hyperlink{cor:intrinsic-necessity}{Corollary 1}} and \textbf{\hyperlink{prop:four-requirements}{Proposition 1}} depend further on E1--E4, especially the exhaustive-partition premise E3 and the premise E4 that at least one intrinsic candidate remains after elimination of the externally enforced class. Readers may reject one or more of these premises, and the conditional structure makes clear where disagreement would lie. If a reader rejects the theorem, the likely point of disagreement is A2 or A3. Even so, the requirement analysis may still clarify what an intrinsic safety-sustaining strategy would need to provide if the paper’s broader framework is accepted. E4 is the decisive existence branch in the argument: if it holds, then at least one intrinsic candidate remains after elimination of the externally enforced class; if it fails, then the paper establishes only the elimination of externally enforced strategies, not the existence of any remaining viable candidates. If a reader accepts the theorem but rejects the necessity result or the four requirements, the likely point of disagreement is E3 or E4. In that case, the paper no longer settles what kind of strategy, if any, could remain; however, the four requirements would still provide guidance for evaluating strategies that claim to be intrinsic, although no longer as necessary conditions.

\textbf{Locality and empirical reach.} \hyperlink{thm:external-impossibility}{Theorem 1} is local in the specific control-theoretic sense used in the proof. It proceeds through a reachable supercritical boundary region \ensuremath{\Gamma} \ensuremath{\subseteq} \ensuremath{\partial}S, not through a global characterization of all dynamics everywhere in state space. Nor does the paper claim that the supercritical regime already holds for current systems. The theorem’s practical applicability therefore depends on whether A2-like boundary conditions arise for the systems and safe sets of interest.

Assessing A2 empirically would require specifying a concrete safe set \emph{S}, identifying boundary regions $\Gamma\subseteq \partial S$ corresponding to safety-critical transitions, and estimating whether system-generated effects near those boundaries can exceed bounded external correction, together with the bound on autonomous drift. In practice, this could involve stress tests, red-team scenarios, deployment simulations, tool-use evaluations, autonomous replication or resource-acquisition tests, and estimates of available correction capacity. These assessments may be especially relevant for agentic, tool-using, or automated-research settings, where system-generated actions can unfold through software environments, laboratories, or deployment infrastructure rather than remain confined to model outputs. Such assessments would not prove A2 universally, but they would make it empirically contestable for particular systems and deployment settings; evidence that available correction capacity can be scaled to consistently stop or reverse such trajectories would weaken A2-like conditions for those systems.

\textbf{Model abstraction and realism gap.} The formal model is deliberately idealized. Actual AI systems are often discrete-time, stochastic, highly non-stationary, and embedded in deployment settings where safety constraints are only partially specified. The present framework abstracts those features to isolate the structural question at issue: whether a claimed safety-sustaining strategy depends on continued external enforcement or instead is sustained by the system’s own internal dynamics. It also treats the safe set \emph{S} as fixed; strategies based on continuously redefining or contracting the safe set are not analyzed here. These idealizations limit the theorem’s direct empirical application, but they do not make the analysis merely hypothetical. Current AI systems already involve bounded external interventions, imperfect monitoring, delayed correction, and increasingly capable endogenous behavior. The formal model isolates the structural pressure created when limited external correction must counteract system-generated effects that may grow in scale, speed, or strategic sophistication. More detailed or extended models incorporating stochasticity, discrete time, or adaptive safe sets may change how this pressure is measured, but not why dependence on continued external enforcement matters.

\textbf{Non-sufficiency.} The four structural requirements are necessary conditions, not a complete solution. They do not show that any existing system, method, or proposal already satisfies them. Their role is to constrain what any candidate safety-sustaining strategy must provide in order to be viable. If additional requirements exist, that would not weaken the paper's conclusion. It would only show that the remaining intrinsic class is even more constrained than the present derivation establishes.

\textbf{Operationalization of \ensuremath{\Phi}}. A current limitation is that \ensuremath{\Phi}, the set of internal configurations corresponding to safety-compatible terminal objectives, is defined conceptually rather than operationally. In contemporary systems, \ensuremath{\Phi} can be understood as the set of internal configurations whose effective decision criteria select actions preserving \emph{S}. These criteria may be implicit in policies, value functions, preference models, tool-use policies, architectural constraints, verifier-augmented decision rules, or other mechanisms. The paper does not show how to identify \ensuremath{\Phi} exactly. Instead, it uses \ensuremath{\Phi} to name the target that any intrinsic safety-sustaining strategy must make precise: which internal configurations are safety-compatible, how they can be detected or approximated, and how they can be maintained under fine-tuning, self-modification-like updates, or capability growth. 

To operationalize \ensuremath{\Phi}, a viable intrinsic strategy must provide a reliable way to check whether the system’s internal configuration remains safety-compatible over time. For current systems, this would likely begin with imperfect proxies rather than exact identification. These could include behavioral evaluations, interpretability evidence, mechanistic-interpretability indicators such as activation-level or circuit-level evidence, and controlled tests of whether the system’s effective decision criteria remain safety-compatible across different tasks, prompts, deployment settings, and capability levels. These proxies would then need to be tested for stability under adversarial pressure, fine-tuning, and other post-training changes. This opens a concrete research program: identifying indicators of \ensuremath{\Phi}-membership and exploring whether improved verification methods could eventually certify safety-compatible internal configurations or their invariance over time.

\textbf{Falsifiability.} The argument is falsifiable in several ways. It would be weakened by a credible demonstration that A2-like supercritical boundary conditions do not arise under development conditions relevant to sustained safety. It would also be weakened by a defensible third structural class of safety-sustaining strategies that does not depend on continued external enforcement yet does not rest on a self-maintaining intrinsic property of the system. Finally, it would be challenged by a viable strategic proposal that avoids the eliminated class while failing one of the four derived requirements. These are substantive points of engagement, not rhetorical escape routes.

\textbf{Open questions.} The most important open questions are now clear. How can \ensuremath{\Phi} be operationalized in real AI systems? What mechanisms, if any, can make terminal-objective invariance robust under self-modification? Can R4, consistency under capability scaling, be verified when system capabilities exceed the conditions under which safety was initially established? Is there any intrinsic architecture or training regime that can credibly satisfy all four derived requirements together? And can A2, the supercritical boundary control-authority gap, be assessed empirically for real systems in concrete deployment settings? The paper does not answer those questions; its contribution is to show that they are unavoidable for any remaining viable safety-sustaining strategy.

\section*{\hypertarget{sec:conclusion}{8. Conclusion}}

This paper establishes two linked conditional results about strategies for sustaining AI safety. The first, formalized by \hyperlink{thm:external-impossibility}{Theorem 1} under explicit premises including a reachability condition, proves a class-wide external impossibility result: once the system’s effects exceed what bounded external control can counteract, no candidate safety-sustaining strategy that depends in any degree on continued external enforcement can sustain safety. The second result, established through \hyperlink{cor:intrinsic-necessity}{Corollary~1} and \hyperlink{prop:four-requirements}{Proposition 1}, shows that if any candidate safety-sustaining strategies remain after that elimination, they must be intrinsic and must satisfy four structural requirements in order to be viable. Safety may not depend on continued external enforcement; the system's terminal objective must be safety-compatible when first formed; that objective must remain stable under self-modification; and safety must continue to be preserved as capability grows. The paper therefore gives formal structure to a widely held concern about the limits of external control: under the stated premises, externally enforced strategies are ruled out as safety-sustaining, and any remaining candidates must be evaluated against the requirements derived here.

\section*{\hypertarget{sec:references}{References}}\begin{paperrefs}

\paperref{amodei2016}{Amodei, D., Olah, C., Steinhardt, J., Christiano, P., Schulman, J., \& Mané, D. (2016). Concrete problems in AI safety. \emph{arXiv preprint arXiv:1606.06565}.}

\paperref{ames2019}{Ames, A. D., Coogan, S., Egerstedt, M., Notomista, G., Sreenath, K., \& Tabuada, P. (2019). Control barrier functions: Theory and applications. In \emph{Proceedings of the 18th European Control Conference} (pp. 3420--3431). IEEE.}

\paperref{askell2022}{Askell, A., Bowman, S. R., Chen, A., Conerly, T., Ganguli, D., Henighan, T., \ldots{} Clark, J. (2022). A general language assistant as a laboratory for alignment. \emph{arXiv preprint arXiv:2112.00861}.}

\paperref{aubin1991}{Aubin, J.-P. (1991). \emph{Viability Theory}. Birkhäuser.}

\paperref{bengio2024}{Bengio, Y., Cohen, M. K., Malkin, N., MacDermott, M., Fornasiere, D., \ldots{} Kaddar, Y. (2024). Can a Bayesian oracle prevent harm from an agent? \emph{arXiv preprint arXiv:2408.05284}.}

\paperref{bhargava2023}{Bhargava, A., Witkowski, C., Looi, S. Z., \& Thomson, M. (2023). What\textquotesingle s the magic word? A control theory of LLM prompting. \emph{arXiv preprint arXiv:2310.04444}.}

\paperref{blanchini1999}{Blanchini, F. (1999). Set invariance in control. \emph{Automatica}, \emph{35}(11), 1747--1767.}

\paperref{bostrom2014}{Bostrom, N. (2014). \emph{Superintelligence: Paths, Dangers, Strategies}. Oxford University Press.}

\paperref{burns2024}{Burns, C., Izmailov, P., Kirchner, J. H., Baker, B., Gao, L., Aschenbrenner, L., \ldots Wu, J. (2024). Weak-to-Strong Generalization: Eliciting Strong Capabilities With Weak Supervision. In \emph{Proceedings of the 41st International Conference on Machine Learning} (Vol. 235, pp. 4971--5012). \emph{Proceedings of Machine Learning Research}.}

\paperref{christiano2017}{Christiano, P. F., Leike, J., Brown, T. B., Martic, M., Legg, S., \& Amodei, D. (2017). Deep reinforcement learning from human preferences. In \emph{Advances in Neural Information Processing Systems} (Vol. 30, pp. 4299--4307).}

\paperref{dalrymple2024}{Dalrymple, D., Skalse, J., Bengio, Y., Russell, S., Tegmark, M., \ldots Tenenbaum, J. (2024). Towards guaranteed safe AI: A framework for ensuring robust and reliable AI systems. \emph{arXiv preprint arXiv:2405.06624}.}

\paperref{everitt2016}{Everitt, T., Filan, D., Daswani, M., \& Hutter, M. (2016). Self-Modification of Policy and Utility Function in Rational Agents. In \emph{Artificial General Intelligence} (AGI 2016) (Lecture Notes in CS, Vol. 9782, pp. 1--11). Springer, Cham.}

\paperref{everitt2018}{Everitt, T., Lea, G., \& Hutter, M. (2018). AGI safety literature review. \emph{arXiv preprint arXiv:1805.01109}.}

\paperref{greenblatt2024}{Greenblatt, R., Shlegeris, B., Sachan, K., \& Roger, F. (2024). AI control: Improving safety despite intentional subversion. In \emph{Proceedings of the 41st International Conference on Machine Learning} (Vol. 235, pp. 16295--16336). \emph{Proceedings of Machine Learning Research}.}

\paperref{hadfieldmenell2016}{Hadfield-Menell, D., Russell, S. J., Abbeel, P., \& Dragan, A. D. (2016). Cooperative inverse reinforcement learning. In \emph{Advances in Neural Information Processing Systems} (Vol. 29, pp. 3909--3917).}

\paperref{hadfieldmenell2017}{Hadfield-Menell, D., Dragan, A. D., Abbeel, P., \& Russell, S. (2017). The Off-Switch Game. In \emph{Proceedings of the Twenty-Sixth International Joint Conference on Artificial Intelligence} (IJCAI-17) (pp. 220--227).}

\paperref{hubinger2019}{Hubinger, E., van Merwijk, C., Mikulik, V., Skalse, J., \& Garrabrant, S. (2019). Risks from learned optimization in advanced machine learning systems. \emph{arXiv preprint arXiv:1906.01820}.}

\paperref{kundu2023}{Kundu, S., Bai, Y., Kadavath, S., Askell, A., Callison-Burch, C., Chen, A., \ldots{} Kaplan, J. (2023). Specific versus general principles for constitutional AI. \emph{arXiv preprint arXiv:2310.13798}.}

\paperref{mazzu2024}{Mazzu, J. M. (2024). Supertrust foundational alignment: Mutual trust must replace permanent control for safe superintelligence. \emph{arXiv preprint arXiv:2407.20208}.}

\paperref{ngo2024}{Ngo, R., Chan, L., \& Mindermann, S. (2024). The alignment problem from a deep learning perspective. In \emph{The Twelfth International Conference on Learning Representations} (ICLR 2024). OpenReview.}

\paperref{nosrati2026}{Nosrati, K., Tepljakov, A., Belikov, J., \& Petlenkov, E. (2026). When control meets large language models: From words to dynamics. \emph{arXiv preprint arXiv:2602.03433}.}

\paperref{omohundro2008}{Omohundro, S. M. (2008). The Basic AI Drives. In P. Wang, B. Goertzel, \& S. Franklin (Eds.), \emph{Artificial General Intelligence 2008: Proceedings of the First AGI Conference} (Frontiers in Artificial Intelligence and Applications, Vol. 171, pp. 483--492). IOS Press.}

\paperref{perrier2025}{Perrier, E. (2025). Out of control: Why alignment needs formal control theory (and an alignment control stack). \emph{arXiv preprint arXiv:2506.17846}.}

\paperref{russell2019}{Russell, S. (2019). \emph{Human Compatible: Artificial Intelligence and the Problem of Control}. Viking.}

\paperref{soares2015}{Soares, N., Fallenstein, B., Yudkowsky, E., \& Armstrong, S. (2015). Corrigibility. In \emph{AAAI Workshop on AI and Ethics}.}

\paperref{tice2026}{Tice, C., Radmard, P., Ratnam, S., Kim, A., Africa, D. D., \& O\textquotesingle Brien, K. (2026). Alignment pretraining: AI discourse causes self-fulfilling (mis)alignment. \emph{arXiv preprint arXiv:2601.10160}.}

\paperref{turner2021}{Turner, A. M., Smith, L., Shah, R., Critch, A., \& Tadepalli, P. (2021). Optimal policies tend to seek power. In \emph{Advances in Neural Information Processing Systems} (Vol. 34, pp. 23063--23074).}

\paperref{yampolskiy2022}{Yampolskiy, R. V. (2022). On the controllability of artificial intelligence: An analysis of limitations. \emph{Journal of Cyber Security and Mobility}, \emph{11}(3), 321--404.}

\paperref{xu2015}{Xu, X., Tabuada, P., Grizzle, J. W., \& Ames, A. D. (2015). Robustness of Control Barrier Functions for Safety Critical Control. \emph{IFAC-PapersOnLine, 48}(27), 54--61.}\end{paperrefs}
\end{document}